# PALM VEIN IDENTIFICATION BASED ON HYBRID FEATURE SELECTION MODEL

Mohammed Hamzah Abed[1], Ali H. Alsaeedi[1], Ali D. Alfoudi[1,2], Abayomi M. Otebolaku[3]
Yasmeen Sajid Razooqi[1]

[1] *University of Al-Qadisiyah / College of Computer Science and Information technology/Iraq*
[2] *Liverpool John Moores University/College of Computer Science/ United Kingdom*
[3] *Department of Computing, Faculty of Science, Technology and Arts, Sheffield, United Kingdom*

*E-mail address: mohammed.abed@qu.edu.iq, ali.alsaeedi@qu.edu.iq, A.S.Alfoudi@2014.ljmu.ac.uk, a.otebolaku@shu.ac.uk , yasameen.sajid@qu.edu.iq, com.post04@qu.edu.iq*

**Abstract:** palm vein identification (PVI) is a modern biometric security technique used for increasing security and authentication systems. The key characteristics of palm vein patterns include, its uniqueness to each individual, unforgettable, non-intrusive and cannot be taken by an unauthorized person. However, the extracted features from the palm vein pattern are huge with high redundancy. In this paper, we propose a combine model of two-Dimensional Discrete Wavelet Transform, Principal Component Analysis (PCA), and Particle Swarm Optimization (PSO) (2D-DWTPP) to enhance prediction of vein palm patterns. The 2D-DWT Extracts features from palm vein images, PCA reduces the redundancy in palm vein features. The system has been trained in selecting high reverent features based on the wrapper model. The PSO feeds wrapper model by an optimal subset of features. The proposed system uses four classifiers as an objective function to determine VPI which include Support Vector Machine (SVM), K Nearest Neighbor (KNN), Decision Tree (DT) and Naïve Bayes (NB). The empirical result shows the proposed system Iit satisfied best results with SVM. The proposed 2D-DWTPP model has been evaluated and the results shown remarkable efficiency in comparison with Alexnet and classifier without feature selection. Experimentally, our model has better accuracy reflected by (98.65) while Alexnet has (63.5) and applied classifier without feature selection has (78.79).

## 1. INTRODUCTION

The unprecedented growth of biometric-based identity management systems pushes governments and private stakeholders to look for more efficient management mechanisms, that has unique physiological or behavioral traits to improve the people's identification and verification[1]. The establishments such as airports or bank transmit very importantly and seriousness data, therefore it requires a strict security system. The traditional security methods based on PIN, magnetic cards, keys and smart cards have become insufficient and week[2]. Moreover, the biometric system appears as a promising solution to improve their security system. A biometric system is a recognition system that allows identifying a certain physiological or behavioural characteristic of individual traits from images (dataset) in order to extract different features[3]. Discrimination of persons using vital factors, includes biometric such as fingerprint, face, iris, palm print, retina, palm vein, hand geometry, voice, signature and gait, has become an urgent necessity to improve biometric security systems. The limitations of using some vital characteristics are significantly suffering from defaulting of extracted features due to disease or damage in their functions[4]. The people who suffer from the disease or have manual work, their fingerprint features might be impossible to extract. In addition, the face-based recognition model still suffers from limitations in terms of stability and lighting. All biometrics could be valid for a dead person or severed human limbs except the palm vein[5]. Dr. K Shumizu[1] introduces the palm vein method, which provides a high possibility of discrimination. Palm vein is a better choice because it is difficult to manipulate or change them. It's features are deep and non-superficial[1], [2], [4], [6]. The palm vein patterns are difficult methods to forge because it is internal in the body and it cannot be valid with the dead parts. Taking into account the aforementioned requirements for palm vein, we propose a hybrid model that consists of 2D wavelet coefficient and particle swarm optimization methods. The hybrid model mechanism aims to ensure the optimization of feature selection of palm vein. The proposed model enhances the prediction of query of machine learning algorithms, where in the proposed system with SVM the classification accuracy data that reaches to 98.65 for left-hand and 98.40 for right-hand.

The rest of this paper is organized as follows. Section 2 presents the related work with palm vein preprocessing, verification and identification. The proposed model and its components are discussed in section 3. The experimental results and analysis are discussed in section 4. Finally, the conclusion of this work and future improvement are presented in section 5.

## 2. RELATED WORK

Recently, in the field of palm vein research, many methods have been proposed for recognition and classification to create a biometric system based on palm vein pattern. All the researchers model mainly consists of three parts: preprocessing, features extraction and classification or recognition phase.

In the paper [1] the researchers proposed a recognition model based on low pass Gaussian filter that aims to use a threshold to convert images into binary image as a preprocessing phase. Moreover, the proposed solution utilized two-dimensional wavelet transform to extract the pattern of palm vein and for image classification, they used the method of cosine minimum distance. In other work [4], the researcher suggested palm vein recognition model based on Gabor filter as a features extraction and proposed set of steps for preprocessing phase starting with adaptive histogram equalization for adjusting the intensity of images, then using Discrete Fourier transform, the coefficient multiply by Gaussian high pass filter. Also, the research used Euclidian distance as matching phase. in paper [5] Al-juboori, Wu and Zhao suggested palm vein verification based on Multiple features, the features that extracted based on local and global features using wavelet and local binary pattern, as well as they used Gaussian matched filter in enhancement part. In the last stage, the researchers used Manhattan distance as a matching method. In other work [7], This paper proposes a multimodal biometric system using vascular patterns of the hand such as finger vein and palm vein images. Initially, the input palm vein and finger vein images are pre-processed make them suitable for further processing. Subsequently, the features from palm and finger vein images are extracted using a modified two-dimensional Gabor filter and a gradient-based technique. These extracted features are matched using the Euclidean distance metric, and they are fused at the score level using the fuzzy logic technique. Research of palm vein verification and region of interest was carried out by Dorota Smorawa [8] the researcher suggests a features extraction methods based on the region of interest to create features vector of palm vein shape geometric. In paper [9] the researcher detects the region of interest based on fast ROI algorithm, then normalize the brightness using adaptive histogram equalization, to prepare palm vein images to be ready for features extraction and classification using Convolutional Neural Network. in the paper [10] the researchers were proposed Otsu thresholding to estimate palm region by binaries the images to segment region of interest, then using 2D Gabor filter for features extraction as well as using Artificial Neural Network for classification and matching phase. Akhmad and his research group in [11] used palm vein for biomedical identification, the palm images are processing by palm boundary detection and Competitive Hand Valley Detection (CHVD) algorithm is used to get reference points for ROI in the cropped image, then used local derivative pattern for features extraction from each pixel of image, in the matching phase two methods are tested histogram intersection and decision making. In the [12] the researcher proposed recognition method based on learning vector Quantization, the images acquisition then preprocessing the images to improve the smoothing of images and detection the region of interest, then from ROI the features extracted using peak point detection of palm, finally the training done by using Learning Vector Quantization. ZHOU and KUMAR in [13] were proposed a model for human identification using palm vein images. The images were preprocessed using contrast stretching and histogram equalization, the enhanced palm vein images were used to extract the features the researcher suggested two methods Neighborhood Matching Radon Transform and Hessian-Phase-Based Feature Extraction

Finally, table 1 shows the summarized of the related works:

TABLE I. RELATED WORKS SUMMARY

| Author | Method description | | |
|---|---|---|---|
| | *Pre-processing* | *Features extraction* | *Matching/ classification* |
| [1]Elnasir et. al | Low pass Gaussian filter | two-dimensional wavelet ransform | cosine minimum distance |
| [2] M.Abed et. al | adaptive histogram equalization, Discrete Fourier transform | Gabor filter | Euclidian distance |
| [5]Al-juboori et. al | Gaussian matched filter | two-dimensional wavelet transform and LBP | Manhattan distance |
| [7] S Bharathi et. al | Enhancement filters | two-dimensional Gabor filter and a gradient-based technique | Euclidean distance and fuzzy logic |
| [8]Dorota | Extract Region of interest | palm vein shapes geometric. | Minimum matching based on point of ROI |
| [9]Nidaa | Fast ROI ,adaptive histogram equalization | CNN | Train Convolutional Neural network |
| [10] Prakash et. al | Otsu thresholding | 2D Gabor filter | Artificial Neural Network |
| [11] Akhmad | boundary detection , CHVD | local derivative pattern | histogram intersection and decision making |
| [12] Herry | Palm boundary and extract RIO | peak point detection | Learning Vector Quantization |
| [13] Zhou et. al | contrast stretching and | Neighborhood Matching Radon Transform and | Matching based on |

| Author | Method description | | |
|---|---|---|---|
| | *Pre-processing* | *Features extraction* | *Matching/ classification* |
| | histogram equalization | Hessian-Phase-Based Feature Extraction | minimum coefficient |

## 3. THE PROPOSED 2D-DWTPP MODEL

The proposed model for palm vein identification model mainly consists of three phases: palm vein images preprocessing, feature extraction, reduction, and classification. Figure 1 illustrates the details of the proposed model.

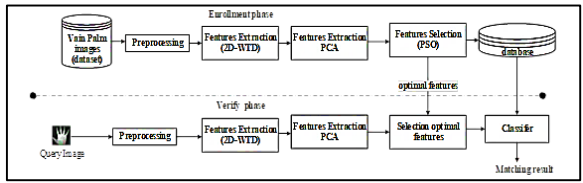

Figure 1. Proposed model structure.

### A. Data Preprocessing

The first phase of our model is a preprocessing of palm vein images. It's used to enhance the images to make the palm's vessel and more visible, the proposed model investigate two steps. The first step, converting the vein image into grayscale enhancement using adaptive histogram equalization to improve the quality of images and make vessels visible. The second step of preprocessing is, converting the images into a negative image, this kind of images show the vessels clearly and it will be used as a final output palm image of the preprocessing phase. Figure 2 illustrates the steps of the preprocessing stage of the palm veinimage.

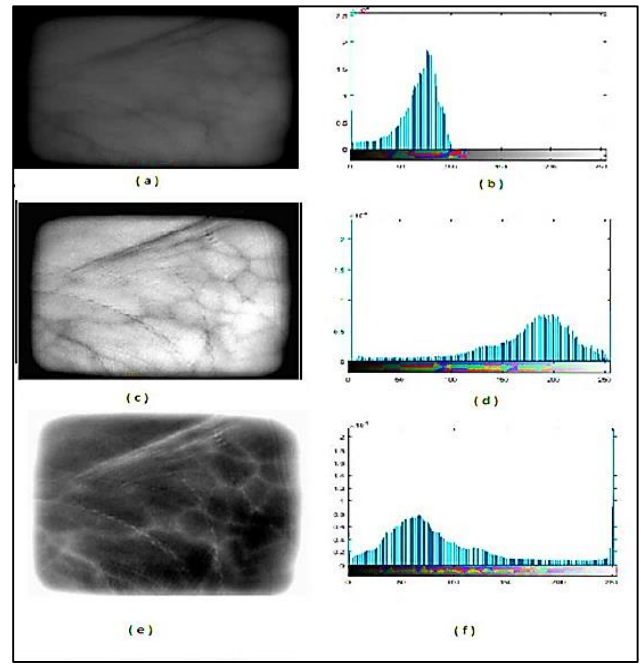

Figure 2. (a) original image (b) histogram of original image (c) adaptive histogram equalization image (d) histogram of adaptive histogram equalization (e) negative image (f) histogram of the negative image

### B. Features Extraction (2D wavelet)

The feature extraction phase aims to extract the relevant features of vein patterns, before moving to the recognition phase. The features are extracted from 2 levels of HARR two-dimensional wavelet coefficient as a raw feature. In order to be able to perform palm vein recognition (PVR), one of the key processes is the extraction of discrimination of the vein patterns. The wavelet functions as an efficient method for representing the image, it's surface and curve[18] .The discrete wavelet transforms (DWT) is a very common technique for analysis of nonstationary signals [14][15][7]. The basic idea of using DWT for representing a signal as a series of low-pass and high pass approximations. The low pass version corresponds to the signal, whereas the high-pass version corresponds to details. This is done at different resolutions. It is almost equivalent to filtering the signal with a bank of bandpass filters whose impulse responses are all roughly given by scaled versions of a mother wavelet[19 ]. Figure 3, shows the two-level composition of an image into four components called 4 sub-images of the wavelet transform. Tow level coefficients of wavelet consider as a raw feature, then by using Principal Component Analysis (PCA) to reduce the number of features and the redundancy, to find a linear set of features that classify into two or more classes of object or event[20]. The main reason for using 2D HAAR DWT for features extraction, because it gives us high information and features of images and all the details of vein vessels.

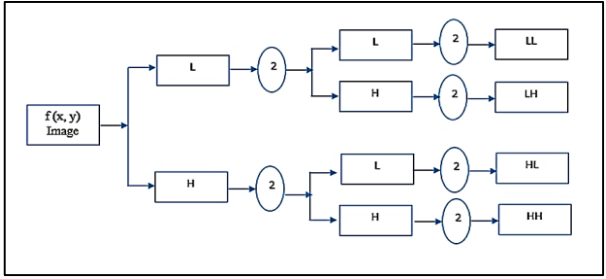

Figure 3. Decomposition of 2D DWT[2]

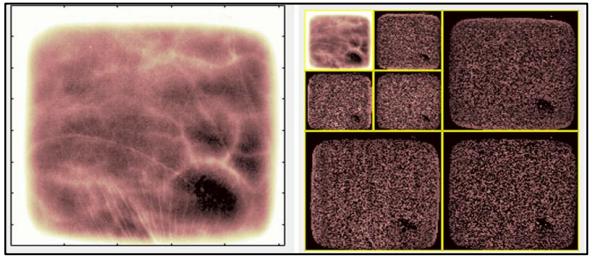

Figure 4. Palm vein

### C. Feature Reduction Based on Principal Component Analysis (PCA)

The extracted features from the palm vein pattern are huge with high redundancy. Furthermore, applied wrapper model on redundancy feature inefficient because it based on selecting features randomly[21], [22]. Therefore, reducing features redundancy is the necessity to enhance the wrapper model. In this paper, the Principle Component Analysis (PCA) is used to feature reduction. It is an orthogonal linear transformation technique that transfers raw data into new form hopefully equal or fewer dimensions of original data [23]. Furthermore, It features reduction and extraction method in data science[24]. Technically, PCA calculates the eigenvectors of a covariance matrix where the highest eigenvalues represent the significant features.

### D. Particle Swarm Optimization (PSO)

PSO is an Evolutionary Algorithms (EA) that was inspired by animals social behavior that live in groups[26]. PSO is preferred over other EA because it has a simple mathematical model with the minimum number of variables. Structurally, it works like the evolutionary optimizers by starting randomly and using all particles to find the optimal solution[27]. The search processing of PSO is technically based on groups of particles. Each particle has its velocity and position, the velocity and position are updated dynamically during search progress. The velocity has been added to the position to generate a new solution. The best solution of PSO's particle is called the local best optimum ($p_{best}$). The global best solution ($g_{best}$) is the best choice among $p_{best}$'s and is updated after reaching the optimizing iteration. Equation 1 computes the velocity of particles[26].

$$V_i^d(t+1) = w(t)V_i^d(t) + c_1 r_1 \left(pbest_i^d - x_i^d(t)\right) + c_2 r_2 (gbest^d - x_i^d(t)) \quad (1)$$

where: $r1$ and $r2$ are random variables in the range [0, 1]. $c1$ and $c2$ are positive constants. $w$ is the inertia weight. $v_i^d(t), x_i^d(t)$ indicate the velocity and position of $i^{th}$ particle at iteration t in $d^{th}$ dimension, respectively. The PSO uses equation 2 to find the new value of particle position (candidate solution)[26].

$$x_i^{t+1} = x_i^t + v_i^{t+1} \quad (2)$$

Where: $x_i^t$ is old particle value, and $x_i^{t+1}$ is a new particle value.

### E. Wrapper Model

Wrapper model is a feature selection method tests deferent groups of features then selects the group that satisfying best result [21]. It has in selecting features based on randomness. The metaheuristic techniques success in applying randomness for search on an optimal solution. Therefore, it significantly optimizing in selecting features that feed into the wrapper model. Figure 5 illustrates the principles of the wrapper model with metaheuristic for selecting the optimum feature.

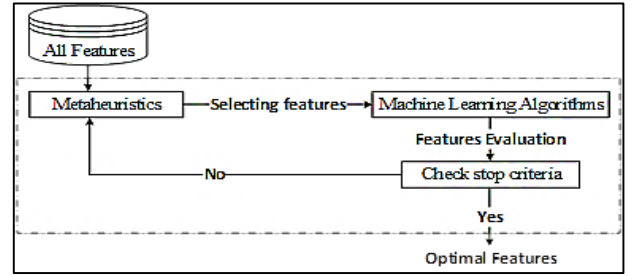

Figure 5. Features Selection based on Wrapper model and Metaheuristic

Generally, the Wrapper model is based on binary feature selection- only features that corresponding 1 in optimization algorithm vector are selected- that restricts the metaheuristic algorithms to search in limited boundaries either 0 or 1[28]. The extending search space is enhanced performance of metaheuristic algorithms[29]. The applied space searching interval range is [-5,+5] in the proposed system. The threshold for feature selection is determined by the user for each experiment[29]. Figure 6 illustrates the encoding and selecting features.

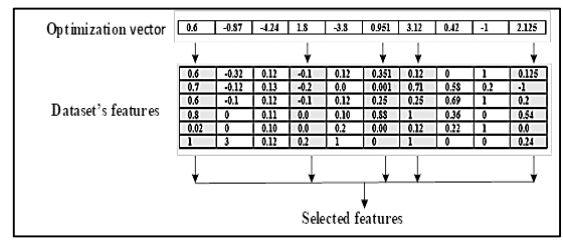

Figure 6. features selecting with threshold 0.5

The wrapper model selects the reverent features based on metaheuristic optimization technique. Technically, It is a greedy optimization algorithm which aims to find the best feature subset group by calculating the accuracy of each group [25]. It repeatedly searches on optimal subset features until it reaches to determinants - stop criteria - of optimization operation [27].

*F. Classification*

In this research four object classifiers has been used as an object function to determine the pattern of palm vein.

- K Nearest Neighbors (KNN)

K nearest neighbors (KNN) is a non-parametric technique used in supervised machine learning to classify new object based on the high-density class of the nearest available cases. Generally, the distance between objects is calculated by one of Minkowski Distances (Manhattan, Euclidean Distance, and Distance), in this work the Euclidean distance(Eq 3) has been applied to find the distance.

$$d_{x,y} = \sqrt{\sum_{i=1}^{n}(x_i - y_i)^2} \quad (3)$$

- Support Vector Machine (SVM)

A Support Vector Machine (SVM) is a discriminative supervised machine learning introduced by xx6x as both regression and classifier model. Technically, it classifies a new object based on Hyperplane and Support Vectors. The hyperplane is multiple lines detected the boundaries of classes that help to determine the class the data objects easily. SVM model set the diminution of hyperplane based on features present in the dataset[22]. Equation 4 calculates the hyperplane

$$w.x + b = 0 \quad (4)$$

Where : $x$ is input, $w$ is weights vector, and $b$ is bias. The dataset that has few numbers of features often is linearly separable, therefore, The SVM uses Eq 5 to classify a new object $x$

$$f(x) = sign(w.x + b) \quad (5)$$

The high dimensional data potentially is not in every case linearly separable as a results the nonlinear decision function (Eq 6) is used to classify a new object $x$

$$f(x) = \text{sign}(\sum_{i=1}^{N} a_i y_i K(x_i \cdot x) + b) \quad (6)$$

Where: $K(x_i \cdot x)$ is the Kernel function, $y_i$ class data.

- **Decision Tree (DT)**

A Decision Tree (DT) is a non-parametric supervised machine learning model that predicts the class of query data based on a sequences series of decision rules[30]. The root of each decision rule is a feature of data has the highest information gained than others. Equation 7 calculates information gain of feature $A$ for dataset $D$.

$$Gain(A) = \sum_{i=1}^{n} A_{p_i} log_2(A_{p_i}) - \sum_{i=1}^{m} d_{p_i} log_2(d_{p_i}) \quad (7)$$

Where $d_{p_i}$ is the probability of each class in the dataset, $A_{p_i}$ probability of each distinct entity in feature $A$

- Naïve Bayes (NB)

Naive Bayes is a probabilistic supervised machine learning model predicts the category of query data based on core concepts Bayes' theorem[31]. Technically, it calculates the probability $P(y/x)$ of query data $x$ with all training classes $y$. The high $P(y/x)$ is determined by the new class of $x$. Equation 8 calculates the class probability[32].

$$P(y|x_1, \ldots, x_n) = \frac{P(x_1|y)P(x_2|y)\ldots P(x_n|y)P(y)}{P(x_1)P(x_2)\ldots P(x_n)} \quad (8)$$

$P(x_1|y)$ is calculated by Gaussian Naive Bayes (Eq9)

$$P(x_i|y) = \frac{1}{\sqrt{2\pi\sigma_y^2}} \exp(-\frac{(x_i - \mu_y)^2}{2\sigma_y^2}) \quad (9)$$

Where $\mu_y$, $\sigma_y$ is mean and standard deviation of train class $y$

**4. EXPERIMENTS AND RESULT ANALYSIS**

We have evaluated the performance of palm vein image classification based on standard dataset PUT Vein Database. Then using 2D HAAR Wavelet 2 level as the feature extraction from the entire image. The experimental test result based on raw features and reduction features using PCA is compared and explaining in the next sections.

*A. PUT Vein Dataset*

The palm vein image dataset that we used in our experiments is PUT Vein Database[3]. It is containing 1200 human vein pattern images for left and right hands; the researcher collected the data from 50 persons for both of hands. The images have been taken in three sessions, four images each time, at least one week between each session. The image in the database has 1280x960 resolution and saved as a 24-bit bitmap

*B. Result and Experimental discussion*

The proposed model is compared with Convolutional Neural Network by using AlexNet as a pre-trained CNN [33]. AlexNet contains majorly eight layers with weight, the first five are convolutional layers and the remaining three are fully connected layer is fed to a 1000-way softmax which produces a distribution over the 1000 class labels. Relu is applied after very convolutional and fully connected layer [2]. Table II illustrates the accuracy of vein palm image classification based on AlexNet

TABLE II. ACCURACY OF PALM VEIN IMAGE CLASSIFICATION BASED ON ALEXNET

| Algorithm | Dataset | Accuracy |
|---|---|---|
| AlexNet | Left hand | 63.50% |
| AlexNet | Right hand | 62.50% |

Four classifiers are applied to predict the palm vein data, KNN, SVM, NB, and DT. Each classifier is applied individually on raw features, reduction features based on PCA, and with feature selection by wrapper model. In KNN, the number of neighbors (K) has been used in evaluating the performance accuracy of the mentioned models and is set by one in order to reduce the noise (increase the probability of close neighbor) of the classifier. The features that extraction by 2D-DWT is high redundancy, therefore, the application of classifiers gave relatively non-high results. In order to prove the accuracy of features selected by PSO, the algorithms run thirty times and the mean of best results are calculated for each algorithm. Figure 7 and 8 shown search progress of PSO particles without and with PCA respectively.

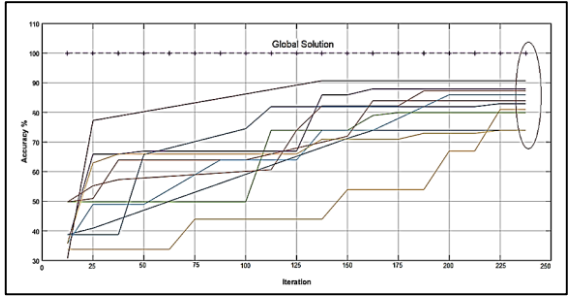

Figure 7. PSO search without PCA (object function is SVM )

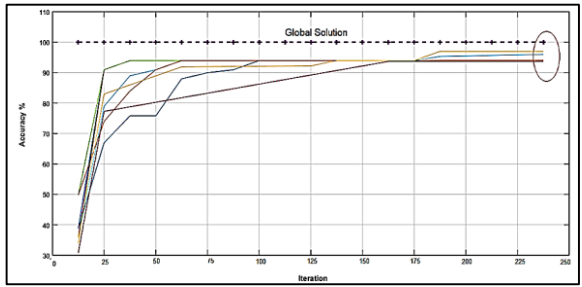

Figure 8. PSO search with PCA (object function is SVM )

The evaluation metrics in term of accuracy is calculated for the four classifiers as shown in tables III.

TABLE III. ACCURACY OF PALM VEIN IMAGE CLASSIFICATION BASED 2DWT.

| dataset | PCA | Feature Selection | Accuracy | | | |
|---|---|---|---|---|---|---|
| | | | KNN | SVM | NB | DT |
| Left hand | No | No | 61.89 | 74.29 | 72.18 | 70.89 |
| Right hand | No | No | 58.63 | 78.79 | 74.41 | 71.65 |
| Left hand | Yes | No | 78.16 | 82.47 | 80.64 | 78.48 |
| Right hand | Yes | No | 80.12 | 83.51 | 80.55 | 77.37 |
| Left hand | No | Yes | 84.13 | 92.85 | 90.74 | 87.04 |
| Right hand | No | Yes | 83.94 | 91.02 | 89.12 | 85.94 |
| Left hand | Yes | Yes | 89.22 | 98.65 | 93.34 | 88.00 |
| Right hand | Yes | Yes | 89.10 | 98.40 | 92.74 | 87.19 |

The SVM achieves the best result compared with another model when reduction features by PCA and selected feature by PSO because both SVM and PCA has same principles in splitting data _ both algorithms build their model based on vectorizing data_ [24 ]. The KNN gets lows to result because it relies on the convergence of data. Clearly, the proposed has significant enhancing on classifier accuracy. In other words, the proposed 2D-DWTPP model has consistently delivered accurate classification results in palm vein image datasets.

and demonstrated its value as a potential alternative to VPI. The proposed system works (2D-DWTPP) to improve the performance of the algorithm automated learning because it processes features by reduction the redundancy in data before selecting features. This allows machine learning algorithms to deal with features that have high precision and relevance with object data.

The proposed 2D-DWTPP has remarkably enhanced on the performance of supervised machine algorithms (SVM, KNN, NB, DT) compering with AlexNet as shown in figure 9.

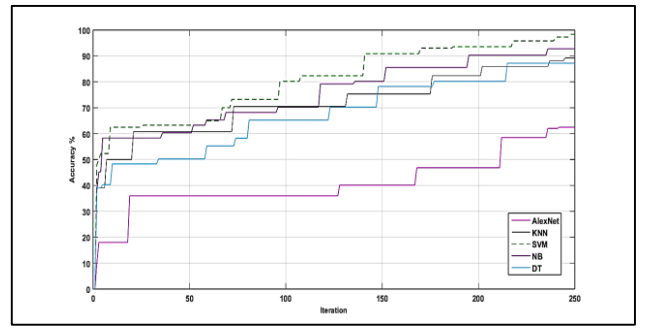

Figure 9. Learning rate of AlexNet and 2D-DWTPP with (SVM, KNN, NB, DT)

The proposed system compared two studies on palm vein identification, Table IV shows the summary of comparison for different techniques on the same dataset in term of accuracy generally, the results show that the performance evaluation of proposed solution outperformed of the other models in table IV, where all the testing cases reflect a significant achievement. In other words, the proposed 2D-DWTPP model has consistently required fewer features to deliver accurate classification results in high-dimensional datasets and demonstrated its value as a potential alternative to traditional features selection.

TABLE IV. SUMMARY OF THE COMPARISON

| Algorithm | Dataset | Accuracy |
|---|---|---|
| Proposed model | PUT vein images | 98.65 |
| Lefkovits [ 34 | PUT vein images | 84 |
| MHAbed [2] | PUT vein images | 94 |

## 5. CONCLUSION AND FEATURE WORK

In this paper, we proposed the Palm Vein verification model using 2D Haar -Discrete Wavelet transform for features extraction, in addition, PCA for features reduction with Particle Swarm Optimization (PSO) for features selection. Moreover, we have used two-level decomposition of wavelet. The accuracy result was compared with the different algorithms of features extraction. From the experiments, the PCA and PSO are significantly improved the performance of classification. The features that extract by 2D-Discrete Wavelet transform have high redundancy. It is not appropriate to implement feature selection algorithms directly because they may choose recurring attributes that in turn negatively affect system performance. Therefore, it is necessary to implement the feature reduction algorithm as well as selection to increase the quality of the produced features. PUT images dataset that used in this model to test the accuracy performance of the proposed model with SVM classifier achieves accuracy was 98.65%. in the future, we are planning to investigate and improve features extraction techniques and feature selection process based on PSO with a minimal computation time and improved recognition accuracy.

.

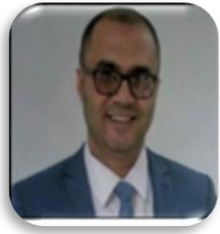

**Ali H. Alsaeedi** is completed B.Sc. in Computer Sciences in 2006 from the college of sciences at University of Al-Qadisiyah, Diwaniya, Iraq. Received his M.Sc. (master) in computer sciences in the year 2016 from the college of computer sciences at the Yildiz Technical University (YTU), Istanbul, Turkey. He has worked as a lecturer at a number of the Iraqi Universities in the areas of Artificial Intelligent, Data mining, and signal processing. He currently works as a lecturer in the University of Al-Qadisiyah. His research interests machine learning, smart optimization algorithms, and optimization of Big Data. Ali has several publications in the areas of the binary of metaheuristic optimization and data mining and a short biography

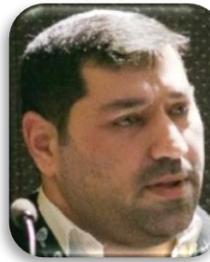

**Ali Saeed D. Alfoudi** received the B.Sc. degree in computer science from the Department of Computer Science, Al-Mansour University College, Baghdad, Iraq, in 2001, the M.Sc. degree from the Department of Computer Science, Fergusson College, University of Pune, India, in 2009, and the Ph.D. degree from Department of Computer Science, Liverpool John Moores University (LJMU), Liverpool, U.K., in 2018. Since 2009, he has been a Lecturer with the Department of Computer Science, University of Al-Qadisiyah, Al Qadisiyah, Iraq. His current research interests include network resource allocation, heterogeneous wireless networks, network virtualization, software-defined networks, cloud computing, mobility management, and the IoT networking.

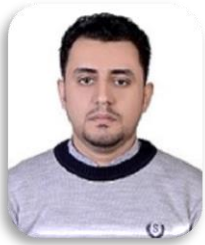

**Mohammed H. Abed** He received the B.Sc. degree in Computer Science from University of Al-Qadisiyah , Iraq 2008, M.Sc. degree in Computer Science from B.A.M. University ,India 2011.currently, he works as a lecturer at the Department of Computer Science, University of Al-Qadisiyah and he doing the researches in the Medical Image Processing and machine learning.

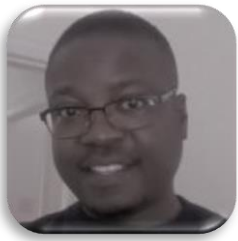

**Abayomi M. Otebolaku**: He received the Ph.D. degree in telecom. engineering from the Faculty of Engineering, University of Porto, Portugal, in 2015. From 2009 to 2015, he was a Research Engineer with the Centre for Telecom. and Multimedia, INESC TEC (formerly INESC Porto). He was a Postdoctoral Research Associate with the Department of Electronics, Telecommunications and Informatics, University of Aveiro, and the Institute of Telecommunications, Aveiro, Portugal. As a Postdoctoral Research Associate, he was with the Department of Computer Science, Faculty of Engineering and Technology, Liverpool John Moores University, Liverpool, U.K. He is currently a Lecturer and a Module Leader in software engineering with the Department of Computing, Sheffield Hallam University, Sheffield, U.K. With participation in several research projects, including European projects, his research focuses on context awareness, activity context recognition, mobile data management, and the IoTdriven personalized services in pervasive and ubiquitous environments. He received the INESC TEC grants for doctoral research, from 2009 to 2011. From 2011 to 2015, he received the Portuguese government

Fundacao para Ciencia e a Tecnologia-Foundation for Science and Technology (FCT) full doctoral grants (Bolsa de Deutoramento)

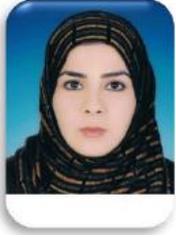 **Yasameen Sajid Razooqi:** she received the B.Sc degree in Computer Science from University of Al-Qadisiyah , Iraq ,with very good grade (top 5 graduated student) 2011. she is works as a teacher of Computer Science subject in brilliant high school for girls in Al-Diwaniya city, Now she is M.Sc Student in College of Computer Science and IT , University of Al-Qadisiyah. and she is doing research based on WSN simulation and routing, IOT, machine learning and artificial Intelligent.